\documentclass[conference]{IEEEtran}
\usepackage{cite}
\usepackage{amsmath,amssymb,amsfonts}
\usepackage{algorithmic}
\usepackage{graphicx}
\usepackage{textcomp}
\usepackage{xcolor}
\usepackage{subcaption}
\usepackage{float}
\usepackage{hyperref}
\usepackage{multirow}
\usepackage{booktabs}
\def\BibTeX{{\rm B\kern-.05em{\sc i\kern-.025em b}\kern-.08em
    T\kern-.1667em\lower.7ex\hbox{E}\kern-.125emX}}
\begin{document}

\title{Data-Free Generative Replay for Class-Incremental Learning on Imbalanced Data}

\author{\IEEEauthorblockN{Sohaib Younis\IEEEauthorrefmark{1},
Bernhard Seeger\IEEEauthorrefmark{2}}
\IEEEauthorblockA{Department of Mathematics and Computer Science,
University of Marburg\\
Marburg, Germany\\
Email: \IEEEauthorrefmark{1}sohaibyounis89@gmail.com,
\IEEEauthorrefmark{2}seeger@mathematik.uni-marburg.de}}

\maketitle
\thispagestyle{empty}

\begin{abstract}
Continual learning is a challenging problem in machine learning, especially for image classification tasks with imbalanced datasets. It becomes even more challenging when it involves learning new classes incrementally. One method for incremental class learning, addressing dataset imbalance, is rehearsal using previously stored data. In rehearsal-based methods, access to previous data is required for either training the classifier or the generator, but it may not be feasible due to storage, legal, or data access constraints. Although there are many rehearsal-free alternatives for class incremental learning, such as parameter or loss regularization, knowledge distillation, and dynamic architectures, they do not consistently achieve good results, especially on imbalanced data. This paper proposes a new approach called Data-Free Generative Replay (DFGR) for class incremental learning, where the generator is trained without access to real data. In addition, DFGR also addresses dataset imbalance in continual learning of an image classifier. Instead of using training data, DFGR trains a generator using mean and variance statistics of batch-norm and feature maps derived from a pre-trained classification model. The results of our experiments demonstrate that DFGR performs significantly better than other data-free methods and reveal the performance impact of specific parameter settings. DFGR achieves up to 88.5\% and 46.6\% accuracy on MNIST and FashionMNIST datasets, respectively. Our code is available at \href{https://github.com/2younis/DFGRID}{https://github.com/2younis/DFGR}
\end{abstract}
\pagestyle{plain}
\section{Introduction}
Recent advances in neural networks and deep learning have been reported to surpass human capabilities in a wide range of individual tasks or similar multiple tasks \cite{russakovsky2015imagenet}. However, these architectures often remain static after training and cannot adapt their behavior over time or learn from new data. They also require a lot of labeled data to be read as input into the network multiple times, thus requiring significant training time. However, data arrives continuously as a stream of data items or batches in many real-world scenarios. Therefore, machine learning models should be able to learn from a data stream and adapt to a changing environment. This ability to acquire new knowledge is known as continual learning or lifelong learning \cite{parisi2019continual}.

Neural networks learn from new data by re-training the network on the entire old and new dataset or by transfer learning only on the new dataset. Transfer learning is a method in which a model trained on a similar domain is partially re-trained on new data. One of the main problems is that when the network tries to learn from new data or tasks, it interferes with the previously learned knowledge, leading to catastrophic forgetting \cite{mccloskey1989catastrophic}, which causes a significant decrease in performance or a complete loss of old knowledge.

To avoid catastrophic forgetting, the learning architecture must acquire new knowledge while preventing it from interfering with old knowledge. This phenomenon is called the stability-plasticity dilemma \cite{mermillod2013stability}, where stability and plasticity refer to how strongly the systems retain learned knowledge and how much the systems can adapt to learn new knowledge, respectively. Too much stability will impede efficient learning from new data, whereas too much plasticity will result in forgotten knowledge previously learned.

There have been many different techniques for overcoming catastrophic forgetting, where the most common strategies belong to one of three main categories, namely architectural, regularization, and rehearsal \cite{parisi2019continual}\cite{kemker2018measuring}\cite{de2021continual}. In an architectural approach, the network tries to accommodate new knowledge by dynamically changing the number of layers or neurons in the network while also keeping some parts of the network static by fixing the weights of some neurons \cite{rusu2016progressive}\cite{zhou2012online}\cite{draelos2017neurogenesis}. In the regularization approach, the plasticity of the network is controlled by imposing some restrictions on the update of the neurons' weights. This approach generally adds an extra adjustable regularization term to the loss function \cite{li2017learning}\cite{wen2018overcoming}. Finally, in the rehearsal approach, old data is replayed to the model mixed with the current data for joint training. In this approach, the network retains old knowledge while also learning new knowledge \cite{shin2017continual}\cite{lopez2017gradient}\cite{kamra2017deep}. There are two ways of replaying old data: rehearsal and pseudo-rehearsal \cite{robins1995catastrophic}. In simple rehearsal, the old data replayed to the model during training is stored in memory. This method becomes inefficient in many real-world settings because of its high storage space consumption \cite{balaji2020effectiveness}, especially in the case of edge devices \cite{pellegrini2021continual}. On the contrary, pseudo-rehearsal avoids storing old data by generating data on demand randomly or from a generative model.
\begin{figure}[htb]
\centerline{\includegraphics[scale=0.38]{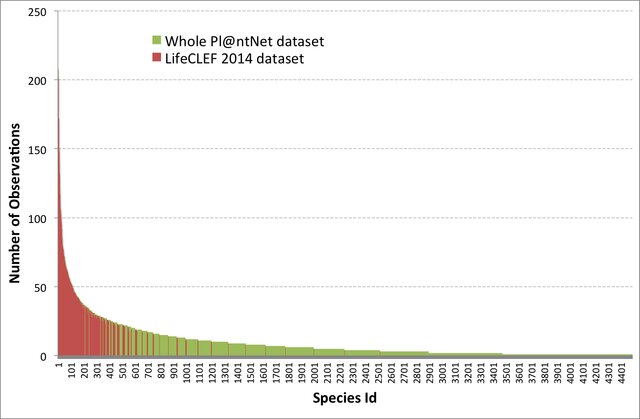}}
\caption{Long tail distribution of the whole Pl@ntNet dataset (with PlantCLEF 2014 subset in red) \cite{joly2014species}.}
\label{fig1}
\end{figure}

In addition to catastrophic forgetting, another challenge of deep learning in a real-world setting is that many image datasets are rarely balanced and follow a long-tail distribution, as shown in Fig.~\ref{fig1}. The plot depicts the frequency distribution of species from the Pl@ntNet data set sorted by the number of occurrences. As shown, the distribution of images in a dataset is not uniform but extremely skewed towards some classes, resulting in a small number of classes containing the majority of images or samples and many classes containing only a small number of images. These class imbalances cause severe learning problems and are an active area of research in machine learning \cite{wang2017learning}\cite{krawczyk2016learning}. The most common approaches to address class imbalance are undersampling of majority classes, oversampling of minority classes, data augmentation, or using synthetic data to balance the classes. In continual learning, an approach similar to oversampling is used mainly, especially in rehearsal or replay-based methods \cite{chrysakis2020online}.

In this paper, we propose a new approach to image classification called Data-Free Generative Replay (DFGR) that addresses the problems of continual class-incremental learning and imbalanced datasets using pseudo-rehearsal from a generator. The generator in DFGR is not trained on real images, as it may not always be available due to privacy concerns or storage limitations. Instead, it uses the means and variances of batch normalization layers \cite{ioffe2015batch} of the trained classifier and the final layer feature maps of each trained class. Since DFGR does not store the data or part of it for replay but solely relies on the pseudo-rehearsal from the generator, it uses focal loss on the classifier to cope with imbalanced data and dynamically adjust the replay probabilities of classes for the generated images. In particular, we present two main contributions in this paper for DFGR: First, a feature map loss for estimating high-level features of the images during the training of the generator, and second, a generator replay adjustment for data augmentation of the generated images.

The remainder of this paper is organized as follows. Section~\ref{background} investigates related works on data-free class incremental learning. Section~\ref{overview} gives an overview of DFGR. Section~\ref{methodology} and Section~\ref{implementation} explain the methodology for DFGR and its implementation, respectively. Finally, the results of an experimental comparison of our approach are shown in Section~\ref{results}.
\section{Background and Related Work} \label{background}
This paper focuses on class incremental learning \cite{rebuffi2017icarl}, one of the three continual learning scenarios presented in \cite{van2019three}. In this scenario, the training model arrives incrementally as a stream of labeled data belonging to different classes. The model accesses the data in stages or episodes, where each stage is called a task. Thus, the model trains on the data in a sequence of tasks, where each task consists of a non-overlapping subset of classes. Each task can consist of one or more classes, but each class can only appear in a single task. The class-incremental learning scenario can be challenging because the model has to learn to discriminate between all classes seen so far, either from current or previous tasks, mainly when they belong to different tasks. This becomes even more challenging in a data-free learning setup where storing any training data from the previous tasks is not allowed or possible, and the only available information is the model trained on previous tasks (with some meta-data) and the training data for the current task.

In the following, we focus on the most recent and effective techniques for class-incremental learning \cite{belouadah2021comprehensive}\cite{masana2022class}. We only cover the methods that support data-free continual learning, whereas replay-based methods are not in the scope of the paper and hence are omitted. We first provide an overview of the methods and then outline the differences in our approach.
\subsection{Regularization-based Methods}
These types of methods avoid storing data previously seen but rely on imposing additional constraints on the update process of various model parameters and hyper-parameters during training to mitigate catastrophic forgetting. Thus, these methods are inherently data-free. There are various options for regularization during training. Some regularize the model weights, \cite{kirkpatrick2017overcoming}\cite{zenke2017continual}\cite{aljundi2018memory}, while others focus on remembering important feature representations of previous data \cite{li2017learning}\cite{dhar2019learning}.

In image classification, one renowned method called Elastic Weight Consolidation (EWC) \cite{kirkpatrick2017overcoming} adds a quadratic penalty to the loss function, which restricts the update of model parameters considered important to the previously learned classes. The importance of the parameters is approximated by the diagonal of the Fisher information matrix \cite{pascanu2013revisiting}. Another similar method is Synaptic Intelligence (SI), which calculates the importance of the learned parameters with the help of synapses \cite{zenke2017continual}. Memory Aware Synapses (MAS) calculate the importance of weights in an unsupervised manner with Hebbian learning by observing the sensitivity of the trained model’s output function \cite{aljundi2018memory}.

Other approaches to regularization aim to prevent activation drift \cite{li2017learning}\cite{dhar2019learning}, which is the change in activations of the old network while learning new tasks. This approach is based on knowledge distillation from a model trained on the previous classes to the model being trained on the new data. A commonly known method with this approach is Learning without Forgetting (LwF) \cite{li2017learning}. It takes the output of the trained model on the new data as the soft labels for previously seen classes and uses them as targets for the new classifier. A recent improvement of this is Learning without Memorizing (LwM), which introduces an attention distillation loss to preserve the attention maps of the classifier on previous classes while training on new data \cite{dhar2019learning}.
\subsection{Knowledge distillation}
Knowledge distillation is a widely used approach to transfer knowledge from a pre-trained model to another model \cite{hinton2015distilling}. Conventionally, most methods using knowledge distillation require access to the previous dataset. However, some recent approaches follow the data-free constraints. Lopes et al. \cite{lopes2017data} attempt to reconstruct the original data from the meta-data (e.g. means and standard deviation of activations from each layer) to reconstruct the original data. Chen et al. \cite{chen2019data} use a pre-trained classifier as a fixed discriminator to train a new generator that could generate images with maximum discriminator responses.

One of the earliest attempts to synthesize images from a trained model without any access to the training data is DeepDream \cite{mordvintsev2015inceptionism}, which generates images from random noise by minimizing classification loss and an image regularization term to steer the generated image away from being unrealistic. DeepInversion \cite{yin2020dreaming} extends DeepDream with a new feature distribution regularization term that minimizes the distance between the feature map statistics and respective batch normalization layers’ (BN) running means and variances. It shows that using the batch normalization based regularization significantly improves the quality of the images. Recent methods employ a combination of losses like cross-entropy, BN alignment, image smoothness, and information entropy to increase diversity in generated images \cite{smith2021always}\cite{xin2021memory}.

Based on these previous methods, we provide a new approach to the scenario of imbalanced datasets. Since regularization methods only use a single model for training and re-training, they are inferior to those that use a generator to reconstruct previous data, hence helping the classifier retain prior knowledge. On the other hand, knowledge distillation methods mostly require a pre-trained teacher model, which is not feasible for our approach because we are re-training the classifier for incremental learning with only a generator model. Therefore, we cannot use these methods \cite{smith2021always}\cite{xin2021memory}. Instead, we use the same data reconstruction and regularization techniques of these methods without any pre-trained teacher model. In the next section, we show how our approach, entitled Data-Free Generative Replay (DFGR), combines all these incremental learning methods and the techniques for mitigating the ramifications of imbalanced training data.
\section{Overview of DFGR} \label{overview}
\begin{figure}
\centering 
\begin{subfigure}[h]{0.4\linewidth}
\centering 
\includegraphics{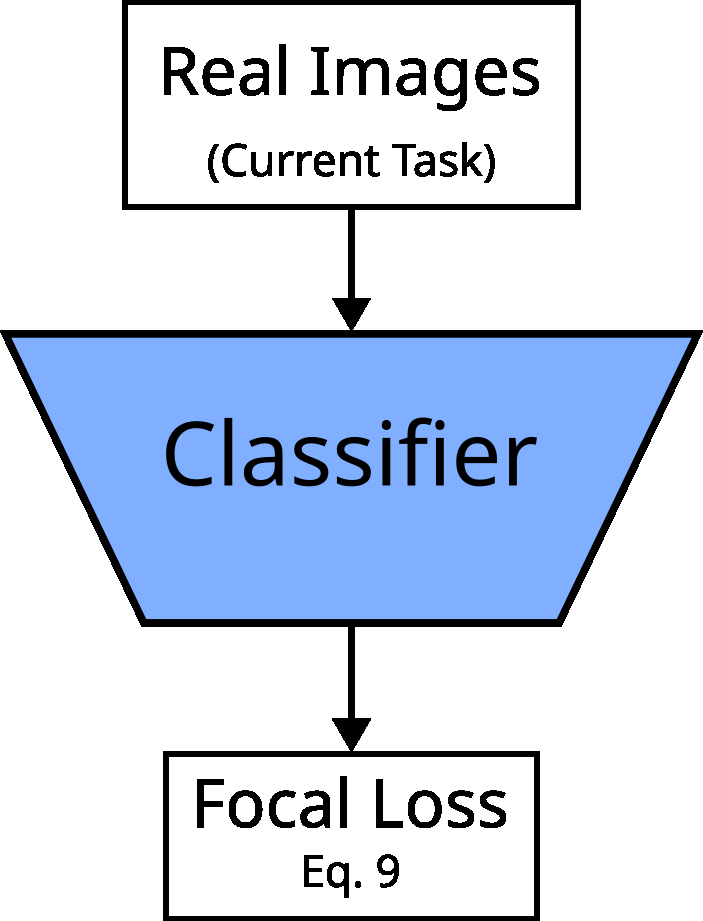}
\caption{}
\label{fig2a}
\end{subfigure}
%\hfill
\begin{subfigure}[h]{0.5\linewidth}
\centering 
\includegraphics{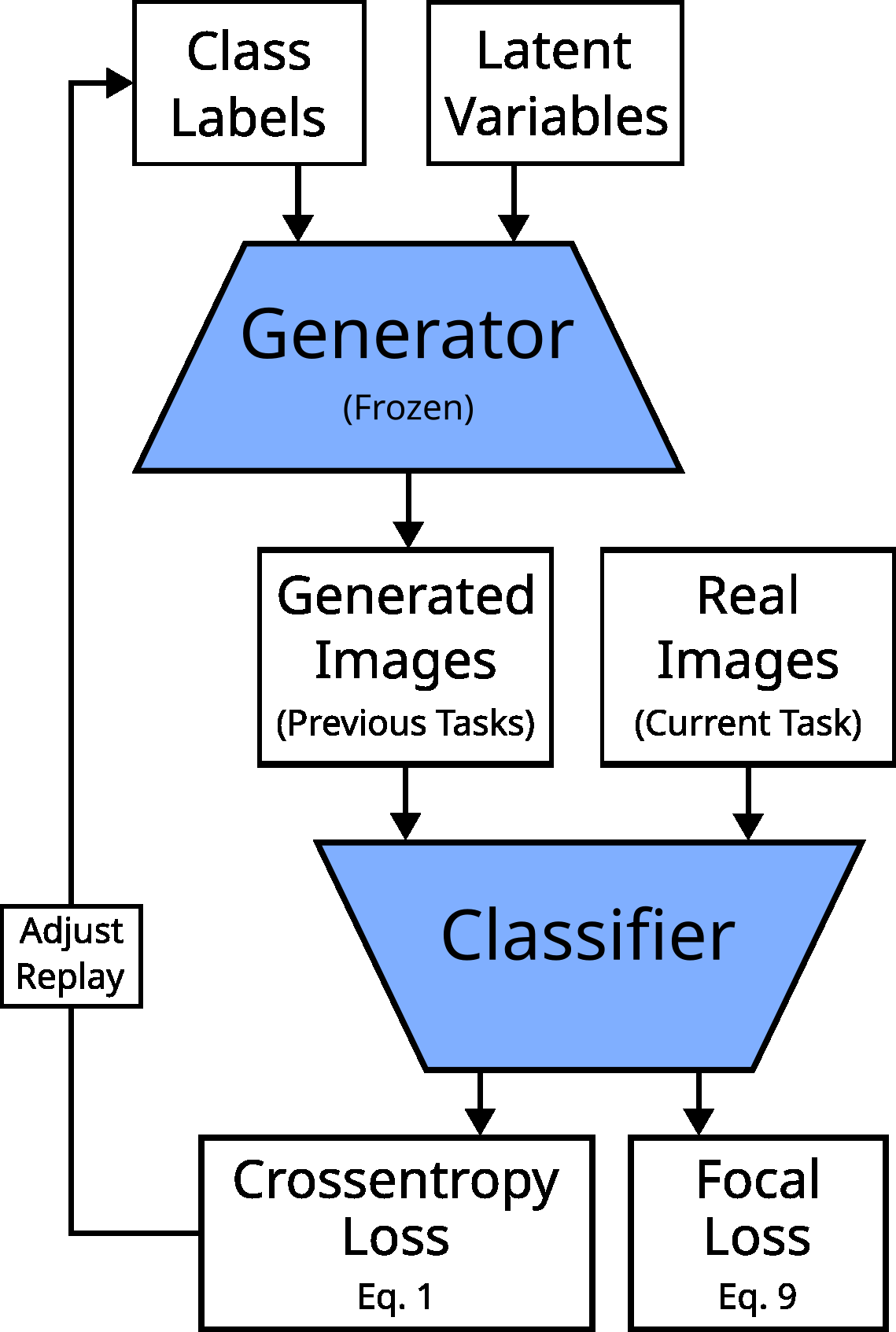}
\caption{}
\label{fig2b}
\end{subfigure}
\caption{a) Workflow of the classifier training with real images (for the first task). b) Workflow for re-training the classifier with real and generated images.}
\label{fig2}
\end{figure}

\begin{figure}[tb]
\centerline{\includegraphics[scale=1]{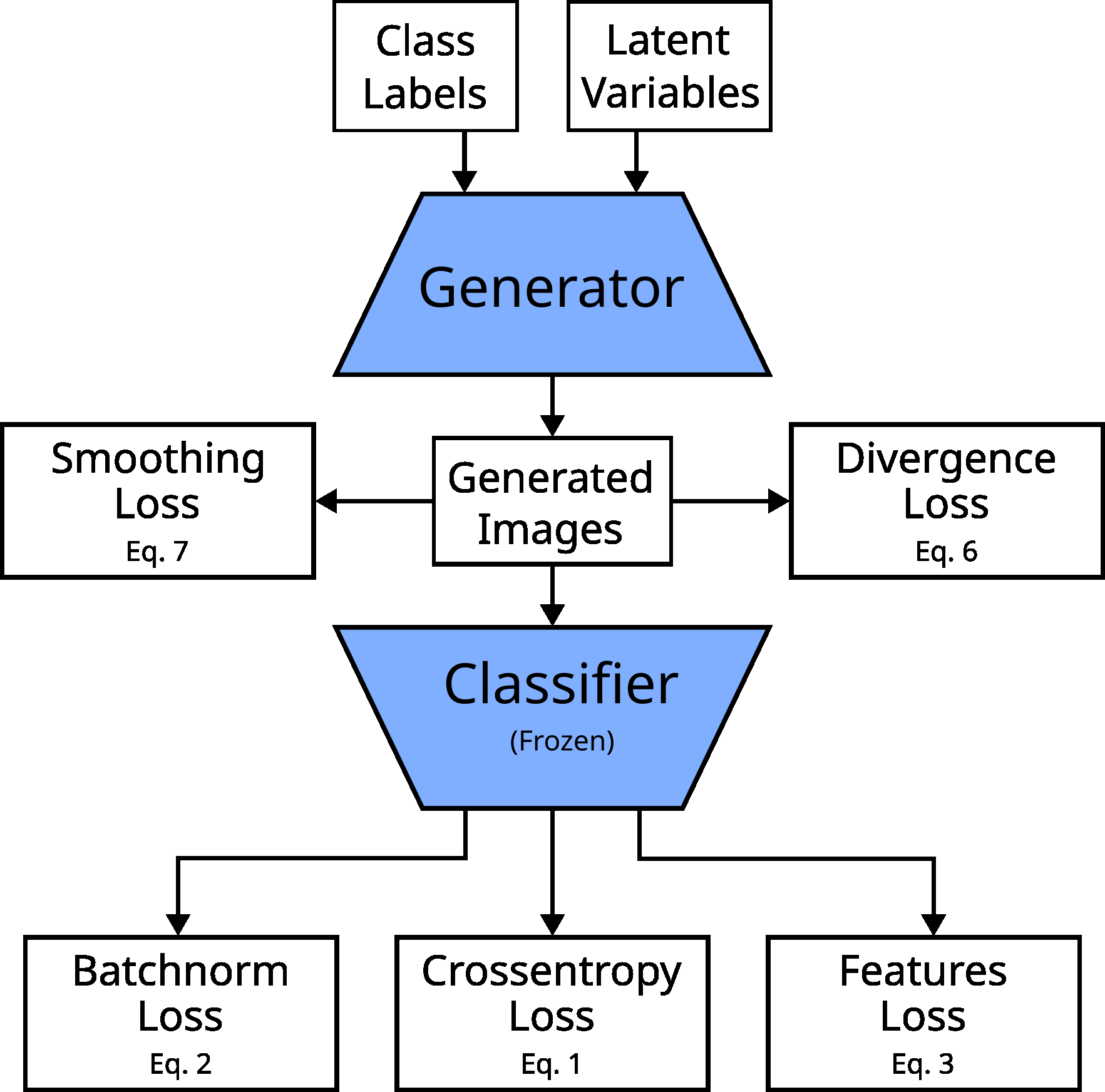}}
\caption{Workflow for training the generator.}
\label{fig3}
\end{figure}

This section gives an overview of the three essential workflows of DFGR and describes its different stages of learning. For each task of the data, the model learns in two sequential stages: 1) Classifier training or re-training, and 2) Generator training. The loss functions used for the classifier and the generator in the workflows shown in Fig.~\ref{fig2} and Fig.~\ref{fig3} contain references (equation numbers) to their definitions detailed in Section~\ref{methodology}.

In the first stage, the classifier is trained on all the available classes, using focal loss\cite{ren2020balanced} instead of cross-entropy loss \cite{goodfellow2016deep}. For training the classifier, the first task uses real data only, while all subsequent tasks use a mix of real and generated data. Fig.~\ref{fig2a} shows the corresponding workflow for the first task. After the first task of data, a mix of the generated images of the previous task and real images of the current task are used to re-train the old classifier. This is called the classifier re-training stage, and Fig.~\ref{fig2b} shows the associated workflow. In this stage, the real images are trained with focal loss as before due to the imbalanced dataset, but the generated images use cross-entropy loss since they are balanced by default. However, the sample ratio for the classes can be adjusted on the fly in case more generated samples are required for a sparsely populated class. This feature of adjusting replay for the generator is one of our main novelties not known from previous work.

Fig.~\ref{fig3} depicts the workflow of the second stage (generator training) where a class-conditional generator \cite{mirza2014conditional} based on BigGAN architecture \cite{brock2018large} is trained to create images similar to the ones of the previous tasks using the trained classifier. The generator can use many different loss functions to achieve the following goals for the generated images: 

\begin{enumerate}
\item An image corresponds to the same classes the classifier has been trained on, Eq.~\eqref{eq2}
\item An image follows a similar data distribution as real ones (i.e.\ small domain gap), Eq.~\eqref{eq3}
\item An image has similar high-level features as real ones, Eq.~\eqref{eq4}
\item There is a sufficient inter-class and intra-class divergence between generated images, Eq.~\eqref{eq7}
\item An image looks realistic with minimum noise, Eq.~\eqref{eq8}
\end{enumerate}
\section{Methodology} \label{methodology}
\subsection{Generator Training}
In the generator training stage, the generator is trained with loss functions, as shown in Fig.~\ref{fig3}, to achieve the five goals for the generated images mentioned in the previous section. We introduce these loss functions in the following subsections. 
\subsubsection{Cross-entropy loss}
Cross-entropy loss is the most common loss function for training a classifier for multi-class classification. We use the Pytorch function for this loss, which implements the negative log-likelihood of the softmax of the logits, given in equation~\eqref{eq2}.
\begin{equation*}
p_{y} =  \dfrac{ exp(z_{y}) }{ \sum_{j} exp(z_{j}) }
\end{equation*}
\begin{equation}
l_{ce} = - log (p_{y})
\label{eq2} 
\end{equation}
As in \cite{goodfellow2016deep}, $z_y$ indicates logits for class $y$, and $p_y$ indicates the softmax probability for class $y$. Thus, $l_{ce}$ is the softmax loss for class $y$. The total softmax loss is the mean of all softmax losses in a batch.
\subsubsection{Batch-normalization statistics loss}
To reduce the data distribution gap between the generated and previous real samples, we introduce a loss that aligns batch normalization (BN) statistics, as used in DeepInversion \cite{yin2020dreaming}. A batch normalization layer in the model keeps the running means and variances of the feature maps of each layer, which are learned during training. These means and variances are used to normalize the input for the next layer, and thus, they reduce internal co-variate shifts of the data \cite{ioffe2015batch}. In particular, they can be utilized to approximate the feature map statistics of real datasets. If the dataset follows a Gaussian distribution, the generated samples should have similar means and variances across the feature maps as the original dataset, whose running means are stored in the pre-trained network. To enforce the similarity of features in all layers, the distance between feature map statistics of real data and generated data should be minimized as expressed in equation~\eqref{eq3}.
\begin{equation}
l_{bn} = \sum_{l} \left\| \mu_{l}(\tilde{x}) - \mu_{l_{bn}}\right\|_{2} + \sum_{l} \left\| \sigma^{2}_{l}(\tilde{x}) - \sigma^{2}_{l_{bn}} \right\|_{2}
\label{eq3} 
\end{equation}
%
% ZZZZ
% 
As in \cite{xin2021memory}, $\tilde{x}$ indicates the feature map vector of the generated data. For layer $l$, $\mu_{l}(\tilde{x})$ and $\sigma^{2}_{l}(\tilde{x})$ denote the estimated batch-wise means and variances of the feature map, respectively, and $\mu_{l_{bn}}$ and $\sigma^{2}_{l_{bn}}$ are the means and variances feature map vectors of the real data stored in the batch normalization layers of the pre-trained network, respectively.
\subsubsection{Feature map loss}
Similar to the batch normalization loss, we can also strive to minimize the distance between features extracted from the last convolutional layer of the model, just before the fully connected layer. Since there is no BN layer between the convolutional network and the fully connected layer, the means and variances of the last layer’s feature map are saved for all the classes after the classifier is trained on all the tasks. During the generator training, equation~\eqref{eq4} is used to minimize the distance between the last feature map’s generated and real data statistics.
\begin{equation}
l_{feat} = \left\| \mu(\hat{x}) - \mu_{m}\right\|_{2} + \left\| \sigma^{2}(\hat{x}) - \sigma^{2}_{m} \right\|_{2}
\label{eq4} 
\end{equation}
In equation~\eqref{eq4}, $\hat{x}$ denotes the convolutional network’s last layer feature maps obtained from the generated data. Moreover, $\mu(\hat{x})$ and $\sigma^{2}(\hat{x})$ are the estimated batch-wise means and variances of the feature maps. Here, $\mu_{m}$ and $\sigma^{2}_{m}$ are the overall means and variances statistics obtained from the saved feature maps of the last layer, depending on the classes in the current batch. $\mu_{m}$ and $\sigma^{2}_{m}$ are calculated for each batch by merging the Gaussian distributions of the classes weighted by the number of samples of each class, hence represented by subscript $m$. Minimizing the feature map loss ensures that the generated images have similar high-level features as the real images.

In the equations~\eqref{eq5} and \eqref{eq6}, $\mu_{c}$ and $\sigma^{2}_{c}$ are the mean and variance of the last layer’s feature maps for class $c$, respectively. Moreover, $n_{c}$ denotes the number of the images in the current batch belonging to class $c$.
\begin{equation}
\mu_{m} = \dfrac{\sum_{c} n_{c} \mu_{c}}{\sum_{c} n_{c}}
\label{eq5} 
\end{equation}
\begin{equation}
\sigma^{2}_{m} = \dfrac{\sum_{c} n_{c} (\sigma^{2}_{c} + \mu^{2}_{c})}{\sum_{c} n_{c}} - \mu^{2}_{m}
\label{eq6} 
\end{equation}
The second main contribution of this paper is the use of feature maps to generate images. Recall that replay adjustment controls the ratio of generated images per class.
\subsubsection{Sample diversification loss} 
Suppose the generator is forced to adjust the BN and feature map statistics of the generated samples to the statistics of the real data. In that case, this can lead to overfitting and reduced image diversity. To increase image diversity within each batch of generated images, we add another loss term as introduced by Xin et al. \cite{xin2021memory}.
\begin{equation}
l_{div} = D_{JS} = -\dfrac{1}{2} \Big( D_{KL}(s_{1} || s_{2}) + D_{KL}(s_{2} || s_{1}) \Big)
\label{eq7} 
\end{equation}
As specified in equation~\eqref{eq7}, the generated images, also called fakes, consist of two samples $s_1$ and $s_2$. They are obtained by first dividing the batch of generated images into two halves and then randomly selecting $2/3$ of the images from each half. The Jensen Shannon divergence ($D_{JS}$) between these two samples is then calculated and used as a loss for maximizing the diversity among generated images.
\subsubsection{Image smoothing loss}
Because generated images can have low-level noise, whereas natural images are generally locally smooth in pixel space, image smoothing loss is introduced as inspired by Smith et al. \cite{smith2021always}. Minimizing image smoothing provides more natural-looking images or, in our case, generated images that look similar to the previously seen real images. Hence, it also lowers other classifier-based losses like BN loss and feature map loss. This loss corresponds to the mean square error between the generated images ($gen$) and a blurred version of the same images by using a Gaussian kernel ($gen_{blurred}$), as shown in equation~\eqref{eq8}.
\begin{equation}
l_{sm} =  \left\| gen - gen_{blurred}\right\|_{2}^{2}
\label{eq8} 
\end{equation}
\subsubsection{Total generator loss}
Finally, the total loss for the generator consists of a linear combination of the individual losses. The hyper-parameters can also be adjusted to activate only some of the individual losses, which we used to assess their impact. The losses that are always activated are image smoothing loss, sample diversification loss, and batch-norm loss. These losses are independent of the class labels of the generated images. The other two losses, cross-entropy loss and feature map loss, depend on the classes of the generated image in each batch. As they are likely to have a high impact in mitigating the effects of imbalanced datasets, we conducted extensive tests with different combinations of these losses to determine their impacts on our approach to balanced and imbalanced data.
\begin{equation}
l_{total} =  \delta l_{ce} + \alpha l_{feat} + \beta l_{bn} + \gamma l_{div} + \epsilon l_{sm}
\label{eq9} 
\end{equation}
In equation~\eqref{eq9}, $\delta$, $\alpha$, $\beta$, $\gamma$ and $\epsilon$ denotes the hyper-parameters of the loss functions. In our experiments, the values of $\beta$, $\gamma$, and $\epsilon$ always equal $1$, and the associated partial sum returns the standard loss as defined in equation~\eqref{eq12}. This standard loss is part of the total loss independent of either the number of images in a class of each batch or the class imbalance of the batch. On the contrary, the cross-entropy loss and feature loss largely depend on the classes of images in each batch. Thus, the values of $\delta$ and $\alpha$ are set to $0$ or $1$ to create different total loss functions and control their contribution to learning imbalanced data.
\subsection{Classifier Training/Re-training}
As the classifier must train on imbalanced data on each task with an unknown distribution, a method is required to make the model learn despite the imbalance in training data. The standard method for image classification is to use cross-entropy loss. However, it can only overcome minor random imbalances per batch, not major imbalances in the whole dataset. To address the imbalance in each class, the training loss for different classes must be reweighed by multiplying them with some weights. We examined focal loss \cite{lin2017focal} and Balanced Softmax loss \cite{ren2020balanced}. Other losses were also considered but were deemed similar in performance. Unfortunately, Balanced Softmax loss could not perform as desired in a class-incremental learning setup, but focal loss, on the other hand, was able to retain some knowledge of previously seen classes.
\subsubsection{Focal loss}
Initially,  He et al. \cite{lin2017focal} introduced focal loss for dense object detection. Most object detection algorithms detect objects of varying size and location by searching millions of bounding boxes per image, which causes class imbalance. Focal loss is a modified version of cross-entropy loss that tries to handle the class imbalance problem by down-weighting easy or abundant classes and focusing on training hard classes. For that, focal loss inversely reweights the classes using their prediction probabilities, as given in equation~\eqref{eq10}.
\begin{equation}
l_{fl} = -(1 - p_{y})^{\gamma} log(p_{y})
\label{eq10} 
\end{equation}
The value of $\gamma$ in the above equation is set to $2$.
\subsubsection{Generator Replay Adjustment}
While training the image classifier on generated images, a generated sample from a particular class can exhibit a higher loss than other classes. We strived to fix this difference by implementing a commonly used technique for data augmentation, repeating or increasing the number of samples for the class with higher loss. As the generated images act as a replay for previously seen data, we monitor the average loss per class and adjust the frequency or probabilities of classes for the next batch of replay images. This adjustment increases the probabilities of classes in each batch with higher average losses, giving them a higher chance of replaying in the next batch.
\subsubsection{Total classifier loss}
The total loss for the classifier is a combination of focal loss for the real data and cross-entropy loss for the generated data, as shown in equation~\eqref{eq11}.
\begin{equation}
l_{total} = (1 - m) l_{fl}^{real} + m l_{ce}^{replay}
\label{eq11} 
\end{equation}
Parameter $m$ is the ratio of the number of the previously seen classes and all classes seen in the current task so far. This parameter prevents the classifier from forgetting the previous classes by increasing their importance depending on their number.
\section{Implementation} \label{implementation}
Our implementation of DFGR is based on a combination of well-established classifier and generator architectures, as shown in Fig.~\ref{fig2} and Fig.~\ref{fig3}. The classifier is derived from the ResNet model \cite{he2016deep}, a renowned deep learning architecture known for its effectiveness in image classification tasks. The generator component uses the BigGAN model \cite{brock2018large}, an advanced conditional generative model capable of generating high-quality images for different classes. We implemented these models in PyTorch framework \cite{paszke2019pytorch}, on a single RTX 3060 GPU.

To test various scenarios and loss functions, we used two well-known benchmark datasets, namely MNIST digits \cite{lecun1998mnist} and FashionMNIST \cite{xiao2017fashion}. These two datasets are chosen due to their relatively small size, making them efficient for experimentation while still being complex enough to evaluate our hypotheses thoroughly. Both MNIST and FashionMNIST contain 10 distinct classes and images are rescaled to 32×32 pixels to match the resolution requirements of our models. Since these datasets are generally balanced, we have introduced some degree of imbalance by selecting a predefined proportion of images per class, ranging from 100\% to 10\% of images for each class. The original datasets served as our baseline, while the newly created imbalanced datasets were used to test our proposed methodology rigorously. Both datasets were assigned to three tasks $T_1, T_2, T_3$ such that every task exclusively contains the images from the following list of classes:

%
% ZZZZ
%
\begin{equation*}
Balanced Data: 
\begin{cases} 
T_1: & \left\{ 3, 4, 9\right\} \\
T_2: & \left\{ 5, 6, 0\right\} \\
T_3: & \left\{ 1, 2, 8, 7\right\} 
\end{cases}
\end{equation*}

For the specification of imbalanced datasets, we added the fraction of total images for each class after the class label (separated by a colon). Again, we examined three tasks $T_1, T_2, T_3$ with the following specifications:

\begin{equation*}
Imbalanced Data:
\begin{cases} 
T_1: & \left\{ 3:1.0, 4:0.6, 9: 0.3\right\} \\
T_2: & \left\{ 5:0.9, 6:0.4, 0:0.2\right\} \\
T_3: & \left\{ 1:0.5, 2:0.7, 8:0.1, 7:0.8\right\} 
\end{cases}
\end{equation*}

For example, task $T_1$ receives 100\%, 60\%, and 30\% of the images from classes 3, 4, and 9, respectively. Thus, we introduce a fixed amount of class imbalance to ensure the repeatability of the experiments.

The hyper-parameters for training our model were selected with careful consideration after checking many different combinations. For ResNet and BigGAN, we chose a batch size of 128 (for training the classifier) and a batch size of 32 (for training the generator), respectively. Both models were trained for maximum 1000 epochs with early stopping, with patience of 50 epochs and 75 epochs for the classifier and generator respectively, ensuring that training terminated when performance improvements stagnated. We utilized the Adam optimizer \cite{kingma2014adam} with a learning rate of $1e-4$ and a $\beta_1$ value of $0.5$ to facilitate model convergence and optimization during training. Because of the long training time required for training the generator and then learning each task sequentially by the classifier, it took around 8 to 12 hours on average for each experiment to run all tasks. Each experiment was repeated at least three times, while most of them around five times.
\section{Results} \label{results}
\begin{table}[h!]
\centering 
\setlength{\tabcolsep}{6pt} % Default value: 6pt
\renewcommand{\arraystretch}{1.1} % Default value: 1
\begin{tabular}{c c c c c}
\toprule
Method&Task I&Task II&Task II\\
\midrule
Class 3&99.9&91.1&87.2\\
Class 4&99.7&91.7&89.9\\
Class 9&99.4&84.8&61.0\\
Class 5&-&99.6&60.0\\
Class 6&-&99.7&91.3\\
Class 0&-&99.9&91.5\\
Class 1&-&-&99.9\\
Class 2&-&-&99.6\\
Class 8&-&-&99.4\\
Class 7&-&-&99.5\\
\midrule
Average&99.7&94.3&88.4\\
\bottomrule
\end{tabular}
\caption{Per class and average accuracy (in \%) of DFGR on MNIST Balanced after training of each task.}
\label{tbl0}
\end{table}
\begin{table*}[h!]
\centering 
\setlength{\tabcolsep}{6pt} % Default value: 6pt
\renewcommand{\arraystretch}{1.1} % Default value: 1
\begin{tabular}{c c c c c c c c c}
\toprule
\multirow{2}{*}{Methods}&\multicolumn{2}{c}{MNIST Balanced}&\multicolumn{2}{c}{MNIST Imbalanced}&\multicolumn{2}{c}{FashionMNIST Balanced}&\multicolumn{2}{c}{FashionMNIST Imbalanced} \\ \cmidrule{2-9}
&Acc.&Avg. Time&Acc.&Avg. Time&Acc.&Avg. Time&Acc.&Avg. Time\\
\midrule
$l_{s}$&45.8&10:38&52.7&8:31&40.0&13:20&39.8&12:23\\
$l_{s} + l_{feat}$&71.7&10:53&58.7&8:49&40.6&13:11&39.7&10:10\\
$l_{s} + l_{ce}$&79.1&11:59&80.5&11:13&43.1&11:54&40.2&11:07\\
$l_{s} + l_{ce} + l_{feat}$&81.7&9:25&81.5&8:09&45.1&11:33&40.3&9:32\\
$l_{s} + ra$&58.0&11:43&53.9&9:43&41.4&10:47&40.0&9:54\\
$l_{s} + l_{feat} + ra$&78.9&9:30&59.7&8:38&43.3&9:51&40.4&7:56\\
$l_{s} + l_{ce} + ra$&87.5&8:45&87.4&8:24&43.6&8:29&42.2&8:31\\
$l_{s} + l_{ce} + l_{feat} + ra$&88.4&8:35&88.5&7:10&46.6&8:40&43.6&8:20\\
\bottomrule
\end{tabular}
\caption{Accuracy (in \%) and average run times (hh:mm) for every dataset, with different combinations of loss functions and replay adjust ($ra$).}
\label{tbl1}
\end{table*}
\begin{table*}[h!]
\centering
\setlength{\tabcolsep}{6pt} % Default value: 6pt
\renewcommand{\arraystretch}{1.1} % Default value: 1
\begin{tabular}{c c c c c c c c c c c}
\toprule
 \multirow{2}{*}{Methods}&\multirow{1}{*}{Models}&\multirow{1}{*}{Model Size}&\multicolumn{2}{c}{MNIST Balanced}&\multicolumn{2}{c}{MNIST Imbalanced}&\multicolumn{2}{c}{FashionMNIST Balanced}&\multicolumn{2}{c}{FashionMNIST Imbalanced} \\ \cmidrule{4-11}
 &\multirow{1}{*}{Saved}&\multirow{1}{*}{(Millions)}&Acc.&Avg. Time&Acc.&Avg. Time&Acc.&Avg. Time&Acc.&Avg. Time\\
\midrule
Naive (Lower Limit)&-&-&41.5&0:34&41.1&0:26&39.9&0:44&39.2&0:22\\
EWC \cite{kirkpatrick2017overcoming}&-&-&47.5&1:23&46.9&0:52&39.9&1:28&39.5&0:58\\
LWF \cite{li2017learning}&Classifier&19.6 M&58.2&1:10&55.5&0:38&41.4&1:08&40.3&0:44\\
\multirow{2}{*}{MFGR \cite{xin2021memory}}&\multirow{1}{*}{Classifier}&\multirow{1}{*}{19.6 M}&\multirow{2}{*}{66.2}&\multirow{2}{*}{15:32}&\multirow{2}{*}{65.8}&\multirow{2}{*}{16:35}&\multirow{2}{*}{42.3}&\multirow{2}{*}{16:05}&\multirow{2}{*}{41.2}&\multirow{2}{*}{16:32}\\
&\multirow{1}{*}{+ Generator}&\multirow{1}{*}{+ 3.2 M}&&&&&&&&\\
\multirow{2}{*}{DFCIL \cite{smith2021always}}&\multirow{1}{*}{Classifier}&\multirow{1}{*}{19.6 M}&\multirow{2}{*}{83.2}&\multirow{2}{*}{13:17}&\multirow{2}{*}{81.1}&\multirow{2}{*}{14:43}&\multirow{2}{*}{48.3}&\multirow{2}{*}{13:27}&\multirow{2}{*}{32.9}&\multirow{2}{*}{14:25}\\
&\multirow{1}{*}{+ Generator}&\multirow{1}{*}{+ 3.2 M}&&&&&&&&\\
\multirow{2}{*}{DFGR (Ours)}&\multirow{1}{*}{Generator}&\multirow{1}{*}{3.2 M}&\multirow{2}{*}{88.4}&\multirow{2}{*}{8:35}&\multirow{2}{*}{88.5}&\multirow{2}{*}{7:10}&\multirow{2}{*}{46.6}&\multirow{2}{*}{8:40}&\multirow{2}{*}{43.6}&\multirow{2}{*}{8:20}\\
&\multirow{1}{*}{+ Features}&\multirow{1}{*}{+ 41 K}&&&&&&&&\\
\bottomrule
\end{tabular}
\caption{Accuracy (in \%) and average run times (hh:mm) for baseline and competitive methods.}
\label{tbl2}
\end{table*}
We conducted a sequel of experiments on MNIST \cite{lecun1998mnist} and FashionMNIST \cite{xiao2017fashion} datasets, considering balanced and imbalanced dataset configurations for each. We examined four different combinations of three loss functions (standard loss $l_s$, cross-entropy loss $l_{ce}$, and feature map loss $l_{feat}$) for class incremental learning, with and without the incorporation of a generator replay adjustment. The standard loss function $l_s$ combines image smoothing loss, sample diversification loss, and batch normalization loss, as shown in equation~\eqref{eq12}.
\begin{equation}
l_{s} = l_{sm} + l_{div} + l_{bn}
\label{eq12} 
\end{equation}
The four combinations of losses are standard loss with cross-entropy loss, standard loss with feature map loss, standard loss with both cross-entropy loss and feature map loss, and finally, just the standard loss. The experiments were conducted using these losses, with and without replay adjustment ($ra$), to test its effectiveness with each loss function. Table~\ref{tbl1} shows the results of these experiments for the four datasets. For each dataset, we report the final accuracy of the classifier on the test set and the average runtime after training on all tasks. For the MNIST Balanced dataset, Table~\ref{tbl0} shows in more detail the development of the average accuracy for each class after the tasks completed their training. Our method DFGR provides a high class accuracy for an increasing number of classes.

As shown in Table~\ref{tbl1}, the combination of all loss functions and replay adjustment gives the highest overall accuracy in our class incremental learning setup, both on balanced and imbalanced datasets. The results show that cross-entropy loss plays the most significant role in improving accuracy, especially when combined with the replay adjustment. Feature map loss also improves accuracy compared to the standard loss, but its impact is more subtle. These findings emphasize the significance of incorporating cross-entropy loss and the advantages of replay adjustment in our method DFGR to enhance the overall performance of class incremental learning models.

To provide a more comprehensive comparison of DFGR, we conducted experiments comparing it against a data-free baseline and two similar methods (MFGR and DFCIL) recently proposed. We tested each method on balanced and imbalanced versions of MNIST and FashionMNIST. Table~\ref{tbl2} shows the final test accuracy and average runtime after training on all tasks. In addition, the table reports the previously stored models required for each method and the model size (number of parameters).

As expected, the naive method showed the lowest accuracy, underlining the necessity for sophisticated incremental learning techniques. Elastic Weight Consolidation (EWC) \cite{kirkpatrick2017overcoming}, a regularization-based approach, performed slightly better than the naive approach. Learning without Forgetting (LwF) \cite{li2017learning}, another regularization-based approach, exhibited improved accuracy compared to EWC, primarily due to its utilization of a classifier trained on previous data to preserve knowledge. In our experimental comparisons, we examined two methods, which, like our method, were developed for data-free learning: Always be dreaming (DFCIL) \cite{smith2021always} and MFGR \cite{xin2021memory}. Both methods employ a classifier trained on previous tasks by a teacher to train the generator and the new classifier on the current task. Consequently, they must store both of these models, significantly increasing the memory requirements for incremental learning.

Our method DFGR emerged as a clear winner in our experiments as it demonstrates higher accuracy than all other methods, as shown in Table~\ref{tbl2}. Among those, only MFGR and DFCIL partly obtain similar results. For one dataset (FashionMNIST Balanced), the accuracy of DFCIL is even slightly better. For FashionMNIST Imbalanced, however, the accuracy of DFGR is more than 10\% higher than DFCIL. In addition, DFGR only requires 15\% of the storage (for the generator parameters and feature map statistics) compared to DFCIL and MFGR. Thus, we conclude that it is not necessary to maintain large classifier models as teachers even if the datasets are balanced. Moreover, these methods also require more time on average for training than DFGR. In summary, DFGR offers a balance between memory efficiency, runtime, and accuracy, making it a promising solution for class incremental learning in resource-constrained data-free environments. It also offers the unique feature to address class imbalance during training.
\section{Conclusion}
This paper proposes DFGR, a novel approach for incremental learning on imbalanced datasets with data-free generative replay. DFGR offers high accuracy without storing previous data, relying solely on a generator trained with a classifier pre-trained on previous data. DFGR is a very appealing method for learning on data streams \cite{gomes2019machine} where historical data is unavailable or only available in a small sliding window. In addition, it could be a candidate for federated learning \cite{li2021survey}, where data might not be accessible for privacy reasons. Future work will also enhance DFGR's performance on larger generic and specialized continual learning datasets, such as ImageNet and CORe50.

\bibliography{manuscript} 
\bibliographystyle{plain}

\end{document}